\documentclass[10pt,letterpaper]{article}  

\usepackage{graphicx}
\usepackage{amssymb}
\usepackage{epstopdf}
\usepackage{graphics}
\usepackage{amsmath, amssymb}
\numberwithin{equation}{section}
\usepackage{graphicx}   
\usepackage{verbatim}   
\usepackage{color}      
\usepackage{subfig}  
\usepackage{hyperref}   
\usepackage{xspace}
\usepackage{float}
\usepackage{amsthm}

\newcommand{\igp}{{IGP}\xspace}
\newcommand{\migp}{{mgIGP}\xspace}

\DeclareMathOperator*{\argmax}{arg\,max}

\newcommand{\bff}{\mathbf{f}}

\newcommand{\bfz}{\mathbf{z}}

\newcommand{\bs}[1]{\boldsymbol #1}

\newcommand{\bfh}{\mathbf{h}}

\newcommand{\hdata}{\bfz^{(h)}_{1:t}}

\newcommand{\peye}{p(\bff^{(i)}\mid\bfz_{1:t}^{(i)})}

\newcommand{\pigpshort}{p(\bff^{(R)},\bff\mid\mathbf z_{1:t})}

\newcommand{\probot}{p(\bff^{(R)}\mid\bfz^{(R)}_{1:t})}
\newcommand{\pperson}{p(\bfh\mid\bfz^{(h)}_{1:t})}

\newcommand{\fa}{p(\bfh,\bff^{(R)},\bff\mid\bfz_{1:t})}
\newcommand{\ft}{p(\bfh,\bff^{(R)}\mid\bfz^{(h)}_{1:t},\bfz^{(R)}_{1:t})}

\title{\LARGE \bf
A Unified Approach to 3 Basic Challenges in Shared Autonomy}
  
\author{Pete Trautman
}

\date{}

\begin{document}

\maketitle
\thispagestyle{empty}
\pagestyle{empty}

\begin{abstract}
\noindent We discuss some of the challenges facing shared autonomy.  In particular, we explore (via the methods of \emph{interacting Gaussian process} models (\igp), \cite{trautmanthesis})
\begin{enumerate} 
\item shared autonomy over unreliable networks,
\item how we can model \emph{individual} human operators (in contrast to the ``average'' of a human operator), and
\item how \igp naturally models and integrates sliding autonomy into the joint human-machine system.
\end{enumerate}
We include a Background Section (Section~\ref{sec:background}) for completeness.
\end{abstract}

\section{Prelude}
\label{sec:prelude}
We begin by recalling the assistive teleoperation model of Section~\ref{sec:background}:
\begin{align}
\label{eq:teleop}
\ft &=\psi(\bfh,\bff^{(R)}) \pperson\probot 
\end{align} 
and we list the meaning of each quantity:
\begin{itemize}
\item $\bff^{(R)}$ is the trajectory of the robot through some state space.  For ground robots, a common choice for the state space is $[x(t), y(t), \theta(t)]$; for air vehicles the state could be 
\[
[x(t), y(t), z(t),roll(t), pitch(t), \theta(t)].
\]  
This trajectory is \emph{a-priori} modeled as a random function distributed according to a Gaussian Process (\cite{gpmlras}), 
\begin{align}
\bff^{(R)}\sim  GP(\bff^{(R)};0,k),
\end{align}
which can be trained offline using input-output examples of the robot's kinematics.

Online measurements of the state of the robot $\mathbf z_{t}^{(R)}$ update the GP to
\begin{align}
\probot=GP(\bff^{(R)};m^{(R)}_{1:t},k^{(R)}_{1:t})
\end{align}
(assuming that the data $\bfz^{(R)}_{1:t-1}$ has already arrived) where $m^{(R)}_{1:t}$ is the new mean and $k^{(R)}_{1:t}$ is the new covariance function of the GP (by ``new'', we mean after incorporation of the new data $\bfz^{(R)}_t$).  

Importantly, this model allows nonparametric probabilistic prediction of the trajectory into the future; that is, 
\begin{align*}
\bff^{(R)} \colon [1:T] \to \mathbb (x, y, \theta)
\end{align*}
where $1\leq t\leq T$.   Indeed, $T$ can be as large as one likes (corresponding to how far into the future one predicts); because $\bff{(R)} \sim \probot$, a continuous measure of the uncertainty $k^{(R)}(t)$ exists, although it grows quite large the further one predicts into the future.

We remark on the following: the structure of $\ft$ is such that the individual robot model $\probot$ changes with each new data point $\bfz^{(R)}_t$; in particular, since GPs are nonparametric models, they have the ability to capture some amount of online nonlinearities (such as motor failures, terrain changes, etc).  The extent to which this is true needs to be explored, however.
\item $\bfh$ is the trajectory of the human operator through some state space.  While the state space of the robot can typically be well characterized with physical models, the state space of the human\footnote{see Section~\ref{sec:learn_human_state} for a more nuanced discussion of learning the human state space.} is not immediately clear.  

As a first step, however, we choose the set of operator commands to the robot to correspond to the state of the human; accordingly, we treat the operator input as measurements $\hdata$ \emph{of the human trajectory through this input space}.  In essence, we are regarding the human state as manifesting via the commands to the robot; extrapolating, we assume that one can predict where the human will go in the ``command space'' (appropriately hedged using probability densities).

As a simple example, if the human is operating a joystick that sends velocity commands $v_x,v_y$ to the robot's actuators, then
\begin{align*}
\bfz^{(h)}_t = (v_x,v_y).
\end{align*}
Furthermore, just as for the robot, the human trajectory is \emph{a-priori} modeled as a random function distributed according to a Gaussian Process (\cite{gpmlras}), 
\begin{align}
\bfh\sim  GP(\bfh;0,k).
\end{align}
New measurements of the state of the human $\mathbf z_{t}^{(h)}$ update the GP to
\begin{align}
\pperson=GP(\bfh;m^{(h)}_{1:t},k^{(h)}_{1:t})
\end{align}
(assuming that the data $\bfz^{(h)}_{1:t-1}$ has already arrived) where $m^{(h)}_{1:t}$ is the new mean and $k^{(h)}_{1:t}$ is the new covariance function of the GP (by ``new'', we mean after incorporation of the new data $\bfz^{(h)}_t$).  

In this way, we can predict what we expect the operator to do using probabilistic inference (just as was done with the robot via $\probot$), thereby enabling joint models of the human-robot team that anticipate future situations (at times $T>t$) by making decisions now, at time $t$.
\item $\psi(\bfh, \bff^{(R)})$ is the interaction function between the human and the robot.  In Section~\ref{sec:background}, a particular choice of this function of this choice is discussed.
\end{itemize}
With this model, control of the remote vehicle is accomplished in a receding horizon fashion: upon receipt of a measurement $\bfz_t = (\bfz^{(h)}_t,\bfz^{(R)}_t)$, the model $\ft$ is updated, and the new navigation protocol is taken to be 
\begin{align*}
(\bfh,\bff^{(R)})_t^* = \argmax_{\bfh,\bff^{(R)}} \left[  \ft \right].
\end{align*}
We then take $\bff^{(R)*}(t+1)$ as the next action in the path (where $t+1$ means the next step of the optimal robot trajectory through the joint human-robot space).  At $t+1$, we receive observations $(\bfz^{(h)}_{t+1},\bfz^{(R)}_{t+1})$, update the distribution to $p(\bfh, \bff^{(R)}\mid\bfz^{(h)}_{1:t+1},\bfz^{(R)}_{1:t+1})$, find the MAP, and choose $\bff^{(R)*}(t+2)$ as the next step.  This process repeats until the human-robot team arrives at the destination.

\section{Shared Autonomy over Unreliable Networks}
\label{sec:networks}
We focus here on a few particular aspects of the control of a remote vehicle over unreliable networks: because operator commands can often arrive late, be dropped, or be otherwise corrupted across an arbitrary network, velocity commands such as $(v_x, v_y)$ cannot be literally interpreted.
\subsection{Laggy networks}
Imagine that the operator is viewing an onboard video feed that is 1 second old, due to communication constraints.  Additionally, imagine that the command $(v_x,v_y)$ takes 1 second to return to the remote vehicle.  Thus, the command received onboard the vehicle is 2 seconds old.  Clearly, this information is stale, and if interpreted by the vehicle literally, could destabilize control.  

However, these commands, while stale, are not devoid of information; we suggest instead that the inputs be treated as measurements $\bfz^{h}_{t-2}$ (if $t$ is in seconds) of the human-machine system.  This suggests that a likelihood be placed on the commands
\begin{align}
p(\bfz^{(h)}_{t} \mid \bfh).
\end{align}
If the current time on the remote vehicle is $t$, then the distribution over the human trajectory gets updated to (assuming all measurements prior to $t-2$ have been received in a timely fashion)
\begin{align*}
p(\bfh \mid \bfz_{1:t-2}) = GP(\bfh;m^{(h)}_{1:t-2},k^{(h)}_{1:t-2}).
\end{align*}
We now update the navigation distribution to
\begin{align*}
p(\bfh,\bff^{(R)}\mid\bfz^{(h)}_{1:t-2},\bfz^{(R)}_{1:t})) = \psi(\bfh, \bff^{(R)})p(\bfh\mid\bfz^{(h)}_{1:t-2})\probot.
\end{align*}
However, because we are modeling \emph{trajectories} of the human, the model can naturally incorporate delayed receipt: the data at $t-2$ informs the distribution $p(\bfh\mid\bfz^{(h)}_{1:t-2})$, but when inference is done at time $t$, additional uncertainty has accumulated.  Thus, when we extract the navigation protocol using
\begin{align}
(\bfh,\bff^{(R)})_t^* = \argmax_{\bfh,\bff^{(R)}} \left[  \psi(\bfh, \bff^{(R)})p(\bfh\mid\bfz^{(h)}_{1:t-2})\probot \right]
\end{align}
the distribution around the human trajectory at time $t$ is less peaked, and so is treated as less informative when evaluating $(\bfh,\bff^{(R)})_t^*$ (which is the actual movement the robot executes at time $t$).
\subsection{Lossy networks}
Lossy networks are treated in an identical manner: indeed, if measurement $\bfz^{(h)}_k$ is missing (where $1<k<t$), then our navigation protocol is still
\begin{align*}
(\bfh,\bff^{(R)})_t^* = \argmax_{\bfh,\bff^{(R)}} \left[  \psi(\bfh, \bff^{(R)})p(\bfh\mid\bfz^{(h)}_{1,2,\ldots, k-1, k+1, \ldots, t})\probot \right].
\end{align*}
Again, the effect on the performance of the system is gradual: as more measurements go missing (or are delayed), the less informative $\pperson$ is, and the more the onboard autonomy is trusted (we thus have a natural formulation of sliding autonomy: see Section~\ref{sec:sliding}).

We emphasize that how well this approach performs is tied strongly to the fidelity of the likelihood function $p(\bfz^{(h)}_{t} \mid \bfh)$ (as is the case with any Bayesian approach); an overconfident measurement model can lead to overly confident human trajectory models, which can place too much weight on incorrect human input.   Under confident models will tend to overtrust the onboard autonomy, thus potentially leading to a robot that does not follow the orders of the operator.  Nevertheless, the presence of an uncertain network forces us to treat operator inputs as probabilistic quantities, rather than deterministic ones.

\section{Generalizing Sliding Autonomy}
\label{sec:sliding}
While methods of sliding autonomy have been explored for a wide variety of tasks (see \cite{sliding_cmu}), the amount of autonomy allocated to the robot (or robots) or the operator (or operators) is typically implemented in a manner \emph{independent} of the human-robot team; that is, some independent estimation algorithm determines how much control each entity receives, and then that number is fed into an algorithm that mixes a weighted combination of each intelligence.  

We argue here that our approach integrates the allocation of autonomy and the actual mixing of the multiple intelligences in a single step.  In particular, we revisit our model of assistive teleoperation
\begin{align*}
\ft &=\psi(\bfh,\bff^{(R)}) \pperson\probot. 
\end{align*} 
This model contains an online model of the human operator $\pperson$ and the robot $\probot$.  In Section~\ref{sec:networks}, we discussed how both the human operator model can respond online to varying network conditions; in Section~\ref{sec:individual} we discussed how both the human and robot models can learn, in an online fashion, peculiarities of the individual operator or individual robot (peculiarities of the individual with respect to general psychological theories or CAD based kinematic models, respectively).

More generally, these individual models maintain a measure of uncertainty about the current state of operator or robot---this uncertainty can naturally be interpreted as proportional to the inverse of how much autonomy should be allocated to each entity. The model thus contains an implicit measure of sliding autonomy, which is a natural artifact of the \igp model.

Perhaps more importantly, however, is that this measure of sliding autonomy is incorporated into the final action in a probabilistic fashion: should the uncertainty become large around either intelligence (due to an unreliable network, uncharacteristic behavior, or any other number of anomalies), then the amount of confidence placed in that intelligence becomes reduced upon blending in the function $\psi(\bfh,\bff^{(R)})$.  Mathematically, as $\pperson$ (or $\probot$) becomes more diffuse, its effect on $\psi$ becomes less pronounced.


One can only extract so much information from any system, and when both distributions become diffuse, the overall information content is very low, and so navigation should start to degrade.  The best we can hope for is a graceful degradation.

Succinctly, the final action taken by the remote vehicle is given by 
\begin{align*}
(\bfh,\bff^{(R)})_t^* = \argmax_{\bfh,\bff^{(R)}} \left[  \ft \right].
\end{align*}
Effectively, sliding autonomy (or blended autonomy) is a natural by-product of our formulation: an implicit measure of blending exists in the individual models, while the uncertainty of those individual models feeds into the interaction function.

\section{Background}
\label{sec:background}
\subsection{Blended Autonomy as an Extension of \migp}
\label{sec:fatheory}
\noindent Current theories of {shared autonomy} are dominated by anecdotal evidence and heuristic guidelines.  In \cite{goodrich-mixed-initiative} the three recognized levels of autonomy are listed: adaptive (the agent adjudicates), adjustable (the supervisor adjudicates), and mixed-initiative (the agent and supervisor ``collaborate to maintain the best perceived level of autonomy''). In \cite{goodrich-human-robot-teams}, human robot collaboration schemas are organized around social, organizational and cultural factors, and in \cite{arkin-ethological} the role of ethological and emotional models in human-robot interaction are examined.  Furthermore, actual implementations are typically designed around need, rather than principle (\cite{murphy-rescue-robotics}): either the remote human operator retains complete control of the robot, or the human operator makes online decisions about the amount of autonomy the robot is given.  

Importantly, the work of \cite{draganrss2012} introduces principled user goal inference and prediction methods, combined with an arbitration step to balance user input and robot intelligence.  However, our approach to shared autonomy as an extension of multiple goal interacting Gaussian processes (\migp) (see \cite{trautmanthesis}) unifies the three steps of \cite{draganrss2012}, thus providing a more straightforward framework in which to understand the fusion of human and machine intelligence.  

We also propose that extending \migp could provide a novel mathematical formulation of shared autonomy (which we call \emph{blended autonomy}).  First, recall the \migp model of \cite{trautmaniros,trautmanicra2013}.  Next, suppose a human operator is controlling the robot from a remote location, so the robot is no longer fully autonomous (we continue the narrative of a robot navigating through a crowd of $n$ individuals $\bff = (\bff^{(1)},\ldots,\bff^{(n)})$).  However, rather than treating the human commands as system interrupts, we wish to understand the continuum of blended autonomy in a mathematical way.  Using the navigation protocol derived using $\pigpshort$ as motivation, we could model the joint human operator-robot \emph{system} as
\begin{align}
\label{eq:sharedigp}
p(\bfh,\bff^{(R)},\bff\mid\bfz_{1:t}) = 
 \frac{\psi(\bfh,\bff^{(R)},\bff)}{Z}  p(\bfh \mid \bfz_{1:t})p(\bff^{(R)} \mid \bfz_{1:t})\prod_{i=1}^n \peye.
\end{align}
where $\bfh$ is the is the human operator's \emph{predicted} interests, modeled with a Gaussian process mixture $p(\bfh \mid \bfz_{1:t})$.   The measurement data is now $\bfz_{1:t} = (\bfz^{\bfh}_{1:t},\bfz^{(R)}_{1:t}, \bfz^{1}_{1:t},\ldots,\bfz^{n}_{1:t})$ where $\bfz^{\bfh}_{1:t}$ are the human operator commands sent from time $1:t$.  Additionally, $\psi(\bfh,\bff^{(R)},\bff)$ is the interaction function between the human operator, robot, and human crowd.  One concrete instantiation of this interaction function is
\begin{align}
\psi(\bfh,\bff^{(R)},\bff) = \psi_{\bfh}(\bfh,\bff^{(R)})\psi_{\bff}(\bff^{(R)},\bff)
\end{align}
where $\psi_{\bff}(\bff^{(R)},\bff)$ is the cooperation function from the model $\pigpshort$ and $\psi_{\bfh}(\bfh,\bff^{(R)})$ is an ``attraction'' model between the operator commands and the robot path.  One possible attraction model is
\begin{align}
\psi_{\bfh}(\bfh,\bff^{(R)}) = \exp\left( (\bfh - \bff^{(R)})^{\top}\bs\Sigma^{-1}(\bfh - \bff^{(R)}) \right).
\end{align}
Thus, the operator's intentionality $\bfh$ and the robot's planned path $\bff^{(R)}$ are \emph{merged}---this formulation of $\psi_{\bfh}(\bfh,\bff^{(R)})$ gives high weight to paths $\bfh$ and $\bff^{(R)}$ that are similar, while the probability of dissimilar paths decreases exponentially.  Bear in mind, however, that $\psi_{\bff}(\bff^{(R)},\bff)$ still gives high weight to paths $\bff$ and $\bff^{(R)}$ that cooperate.  All of this is balanced against the (predicted) individual intentionality encoded in the Gaussian process mixtures $\peye$.

As with \migp, the model $\fa$ suggests a natural way to interpret blended autonomy (or blended decision making): at time $t$, find the MAP assignment for the posterior
\begin{align}
\label{eq:sharednav}
(\bfh,\bff^{(R)},\bff)^* = \argmax_{\bfh,\bff^{(R)},\bff}\fa,
\end{align}
and then take $\bff^{(R)*}(t+1)$ as the next robot action.  As new measurements arrive, compute a new plan by recalculating the MAP of the blended autonomy density.  By choosing to interpret navigation under the model 
\begin{align}
\fa,
\end{align}
\emph{blended autonomy} in complex environments is modeled in a transparent way: human commands are statistically weighted against machine intelligence in a receding horizon framework.   

The key insight is that by modeling the \emph{joint} human-robot system, we can blend human and robot capabilities in a single step to produce a superior system level decision.  When the human system and the robot system are modeled independently, it becomes unclear how to fuse the complementary proficiencies of the human and robot agents.

\bibliographystyle{apalike}
{\footnotesize
\bibliography{standard_bibliography}
}

\end{document}